%% file: main.tex
\documentclass[10pt,twocolumn,letterpaper]{article}

\usepackage[pagenumbers]{cvpr}

\input{preamble}

\definecolor{cvprblue}{rgb}{0.21,0.49,0.74}
\usepackage[pagebackref,breaklinks,colorlinks,citecolor=cvprblue]{hyperref}

\def\paperID{47}
\def\confName{3DV\xspace}
\def\confYear{2026\xspace}

\newcommand{\methodname}{GMT\xspace}

\title{\methodname: Goal-Conditioned Multimodal Transformer for 6-DOF Object Trajectory Synthesis in 3D Scenes}

\author{Huajian Zeng$^{1,4,*}$
\quad
Abhishek Saroha$^{1,2,*}$
\quad
Daniel Cremers$^{1,2}$
\quad
Xi Wang$^{1,2,3}$
\vspace{0.3em}\\  $^{1~}$TU München \qquad $^{2~}$MCML \qquad $^{3~}$ETH Zürich \qquad $^{4~}$MBZUAI
}

\begin{document}

\maketitle
\makeatletter{\renewcommand*{\@makefnmark}{}
\footnotetext{* These authors contributed equally.}\makeatother}
\input{sec/0_abstract}

\input{sec/1_intro}

\input{sec/2_related_work}
\input{sec/3_method}
\input{sec/4_experiment}

\input{sec/5_conclusion}
\newpage
{
    \small
    \bibliographystyle{ieeenat_fullname}
    \bibliography{main}
}

\input{sec/X_suppl}

\end{document}

%% file: preamble.tex
\usepackage[dvipsnames]{xcolor}

%% file: sec/0_abstract.tex
\begin{abstract}
  Synthesizing controllable 6-DOF object manipulation trajectories in 3D environments is essential for enabling robots to interact with complex scenes, yet remains challenging due to the need for accurate spatial reasoning, physical feasibility, and multimodal scene understanding. Existing approaches often rely on 2D or partial 3D representations, limiting their ability to capture full scene geometry and constraining trajectory precision. We present \methodname, a multimodal transformer framework that generates realistic and goal-directed object trajectories by jointly leveraging 3D bounding box geometry, point cloud context, semantic object categories, and target end poses. The model represents trajectories as continuous 6-DOF pose sequences and employs a tailored conditioning strategy that fuses geometric, semantic, contextual, and goal-oriented information. Extensive experiments on synthetic and real-world benchmarks demonstrate that \methodname outperforms state-of-the-art human motion and human-object interaction baselines, such as CHOIS and GIMO, achieving substantial gains in spatial accuracy and orientation control. Our method establishes a new benchmark for learning-based manipulation planning and shows strong generalization to diverse objects and cluttered 3D environments. Project page: \url{https://huajian-zeng.github.io/projects/gmt/}.
\end{abstract}

%% file: sec/1_intro.tex
\section{Introduction}
\label{sec:intro}

Generating realistic and controllable 6-DOF object manipulation trajectories in 3D environments is a central challenge in robotics and computer vision~\cite{billard2019trends, siciliano2016springer}. In manipulation tasks, the object trajectory is often closely aligned with the end-effector trajectory of the robot. Given such a trajectory in Cartesian space, inverse kinematics (IK)~\cite{buss2004introduction} can be used to compute the corresponding joint configurations, thereby converting the end-effector path into a full sequence of robot arm motions. This trajectory-centric formulation decouples perception from control~\cite{ratliff2009chomp, ichter2018learning}, allowing flexible integration of downstream planners or controllers and facilitating generalization across tasks and platforms.

However, synthesizing such trajectories in cluttered 3D scenes remains challenging.
\begin{figure}[t]
  \vspace*{-\baselineskip} 
  \centering
  \includegraphics[width=1\linewidth]{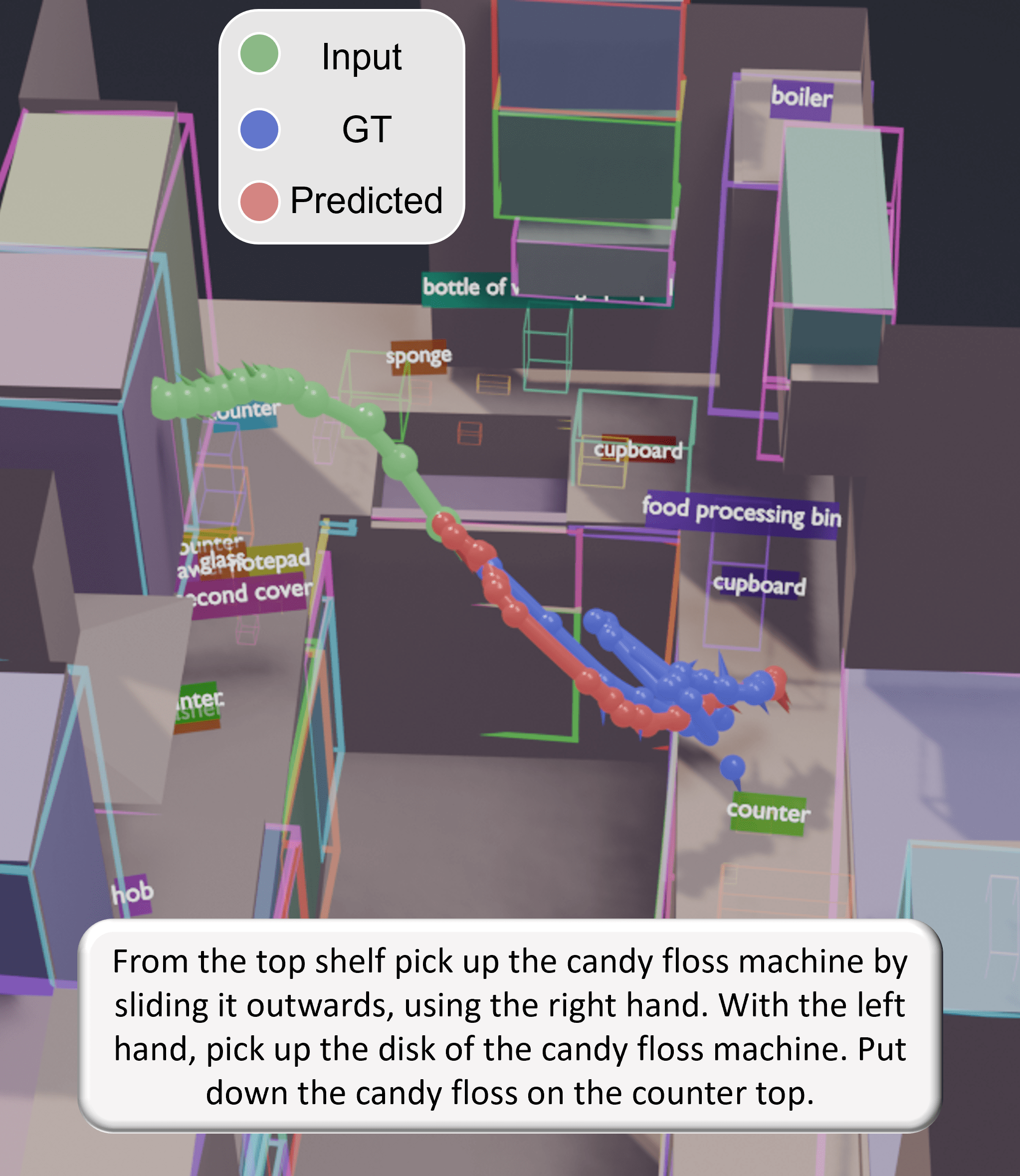}
  \vspace{-1em}
  \caption{Given an observed trajectory, scene context, and action description, our model predicts plausible future 6-DOF object trajectories. The generated trajectories are more efficient than natural human motions.}
    \label{fig:teaser}
    \vspace{-1em}
\end{figure}
First, accurate 3D perception is difficult: depth sensors suffer from noise, occlusion, and sparsity~\cite{yeshwanth2023scannet++, qi2017pointnet++,xu2023mononerd}, where object centers may lie far from any captured surface points~\cite{qi2019deep}.
Second, generated trajectories must respect spatial constraints and physical plausibility, avoiding collisions, maintaining stability, and aligning with object affordances~\cite{gibson1977theory, borja2022affordance, wulkop2025learning}.
Third, goal-conditioned generation requires integrating geometry, semantics, context, and target poses. Traditional planners~\cite{kuffner2000rrt, kalakrishnan2011stomp} face high computational cost in high-dimensional spaces, while most learning-based approaches~\cite{levine2016end, brohan2022rt} predict low-level actions end-to-end, limiting explicit trajectory control or physical constraint injection. 

These challenges highlight the need for generative models that can capture long-range dependencies,
integrate heterogeneous modalities, and enforce structured constraints during synthesis.
Recent advances in transformers~\cite{vaswani2017attention} and diffusion-based generative models~\cite{ho2020denoising, zhang2024motiondiffuse} have demonstrated precisely these capabilities, excelling at modeling complex spatial-temporal structures in high-dimensional spaces.
Nevertheless, applications have focused primarily on human motion~\cite{zhang2023generating, guo2024momask} or human-object interaction. In particular, Human-Object Interaction (HOI) models typically treat objects as passive entities, with object trajectories induced indirectly by human or hand motion~\cite{li2024controllable, gartner2022differentiable, zeng2026flowhoi}.
These approaches~\cite{xu2025intermimic, luo2024omnigrasp} inject HOI behaviors into simulators and rely on reinforcement learning to refine them into executable policies. While effective for human-centric skill transfer, this pipeline restricts the generative model's flexibility and hinders cross-embodiment generalization: the learned policy is tied to specific morphologies and training simulators rather than abstract object dynamics.

\textbf{Our work takes a different perspective}: we shift the focus from human-centric HOI to \emph{object-centric trajectory generation}. By directly modeling 6-DOF object trajectories, we treat objects as primary dynamic entities conditioned on scene and goal constraints. This design allows generated trajectories to serve as a universal intermediate representation: through IK, they can be instantiated by arbitrary robotic embodiments, enabling cross-platform transfer without simulator-dependent policy learning.

In this work, we address these gaps with a multimodal transformer framework for controllable 6-DOF object trajectory synthesis in 3D scenes. Our model jointly leverages geometric, semantic, contextual, and goal information to produce spatially consistent and physically plausible trajectories that can be directly executed on robotic systems via IK.

\noindent In summary, our main contributions are:
\begin{itemize}
    \item \textbf{\methodname, a multimodal transformer architecture} for 6-DOF object trajectory generation, unifying scene geometry, semantics, and task goals within a single framework.
    \item \textbf{A tailored fusion strategy} integrating: (i) geometric conditioning via feature propagation from scene point clouds to 3D bounding box corners; (ii) semantic conditioning via hierarchical category embeddings~\cite{radford2021learning}; (iii) contextual conditioning through global scene features; and (iv) goal conditioning via learnable end-pose embeddings.
    \item \textbf{Extensive experiments} on challenging 3D manipulation benchmarks, achieving state-of-the-art performance over strong human motion baselines such as CHOIS~\cite{li2024controllable} and GIMO~\cite{zheng2022gimo}, with substantial gains in spatial accuracy and orientation control.
\end{itemize}

%% file: sec/2_related_work.tex
\section{Related Work}
\label{sec:related}

Object trajectory synthesis lies at the intersection of computer vision, motion modeling, and 3D scene understanding. While these areas have achieved notable progress, synthesizing controllable 6-DOF object motion in cluttered environments remains underexplored. Below we review related directions and position our work accordingly.

\subsection{Video Prediction \& Dynamics Learning}
Video prediction models aim to forecast object dynamics directly in pixel space. Early methods such as PredNet~\cite{lotter2016deep} and ConvLSTM~\cite{shi2015convolutional} learned short-term temporal dependencies, while Interaction Networks~\cite{battaglia2016interaction} and Visual Interaction Networks~\cite{watters2017visual} introduced relational reasoning between objects. More recent efforts leverage transformers for long-horizon forecasting~\cite{weissenborn2019scaling,wu2021greedy} or diffusion models for stochastic video generation~\cite{ho2022video,harvey2022flexible}. 

These works highlight the importance of modeling dynamics but operate in image space, which struggles with depth ambiguity, occlusion, and 3D consistency. Video generative models such as Sora~\cite{openai2024sora, zhu2024sora} achieve impressive visual fidelity and exhibit emergent properties like object permanence, yet they often lack explicit physical understanding and fail to support planning or decision-making. Similarly, frameworks treating videos as "world models" are hindered by the absence of explicit state-action structure and limited controllability~\cite{li2025worldmodelbench, xing2025critiques}. In contrast, we generate explicit 6-DOF object trajectories in 3D space, enabling precise control over motion and direct interaction with the environment. Our approach focuses on synthesizing physically plausible trajectories that respect spatial constraints, rather than predicting pixel-level dynamics.

\subsection{Human Motion \& Interaction Synthesis}
Human motion synthesis has advanced rapidly, spanning text-conditioned generation~\cite{guo2022generating, tevet2022human}, scene-aware prediction~\cite{zheng2022gimo}, and diffusion-based motion priors~\cite{zhang2023generating, guo2024momask}. Human-object interaction models further integrate semantics and contact reasoning: CHOIS~\cite{li2024controllable} generates synchronized HOI from language prompts, while differentiable simulation~\cite{gartner2022differentiable} enforces physical plausibility. 

Recently, diffusion-based methods have advanced HOI synthesis: CG-HOI~\cite{diller2024cg} explicitly models human-object contact in a joint diffusion framework, improving physical coherence; InterDiff~\cite{xu2023interdiff} introduces physics-informed correction within diffusion steps for long-term HOI predictions; HOI-Diff~\cite{peng2025hoi} utilizes a dual-branch diffusion model plus affordance correction to generate diverse yet coherent human-object motions from text prompts. FlowHOI~\cite{zeng2026flowhoi} proposes a two-stage conditional flow matching framework that generates hand-object interaction sequences conditioned on language instructions and 3D scene context, targeting dexterous robot manipulation by decoupling geometry-centric grasping from semantics-centric manipulation.

Despite these strengths, all of these approaches remain fundamentally human-centric—modeling object motion only as a response to human behavior. In contrast, our work shifts the focus to object-centric trajectory generation, treating objects as primary dynamic entities conditioned on scene and goal constraints. This enables trajectories to be executed via inverse kinematics across robots of varying morphology, rather than being limited to humanoid embodiments.

\subsection{Scene Understanding \& Geometric Reasoning}
Effective motion synthesis requires efficient scene representation. Point cloud methods provide detailed geometry but impose computational constraints. PointNet++~\cite{qi2017pointnet++} addresses some limitations through hierarchical
feature learning on point sets in metric spaces, but still faces computational challenges in dense environments. Voxel representations~\cite{zhou2018voxelnet,maturana2015voxnet} improve efficiency but sacrifice resolution needed for precise
manipulation. Recent fully sparse approaches like VoxelNeXt~\cite{chen2023voxelnext} eliminate sparse-to-dense conversion requirements while maintaining detection performance.

Recent work suggests that coarser representations can be sufficient for many tasks~\cite{deng2025sketchy}. 3D-BoNet~\cite{yang2019learning} demonstrates that direct bounding box regression can be more computationally efficient than
existing approaches by eliminating post-processing steps such as non-maximum suppression, feature sampling, and clustering. This key insight, that high fidelity is not always necessary, suggests that bounding boxes provide sufficient
geometric information for trajectory synthesis while enabling real-time performance.

Multimodal fusion architectures enable flexible combination of geometric and semantic information. Perceiver~\cite{jaegle2021perceiver} provides a scalable blueprint, while Perceiver IO~\cite{jaegle2021perceiverio} extends this with flexible querying mechanisms for structured inputs and outputs. SUGAR~\cite{chen2024sugar} demonstrates effective multimodal pre-training for robotics through joint cross-modal knowledge distillation. The key insight from robotics
applications~\cite{shridhar2023perceiver} is that fusion must respect constraint hierarchies: hard geometric constraints should dominate soft semantic preferences to ensure physically valid output. Our framework builds on these insights by using 3D bounding boxes as a compact yet expressive representation, and enforcing a fusion hierarchy where hard geometric constraints dominate semantic cues.

Furthermore, spatial reasoning in cluttered environments increasingly benefits from hybrid symbolic geometric approaches, where discrete scene graphs capture semantic relations while continuous modules preserve metric precision~\cite{johnson2018image}. This dual representation allows agents to reason over both affordances and spatial feasibility, bridging perception and action. 

%% file: sec/3_method.tex
\section{Methodology}
\label{sec:method}

\begin{figure*}[t]
    \centering
    \includegraphics[width=1\textwidth]{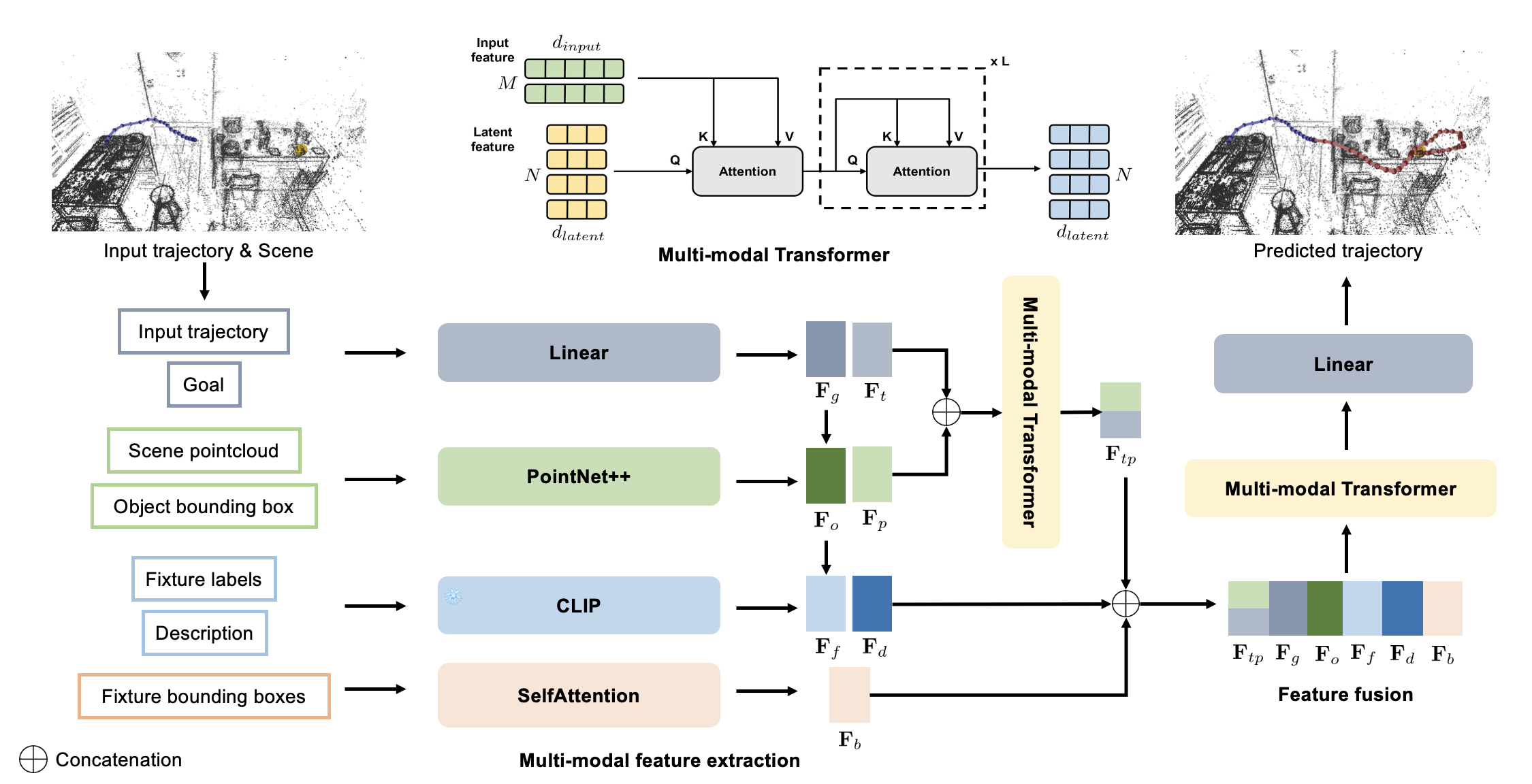}
    \caption{
    \textbf{Pipeline overview.}
    Given an observed trajectory and scene context, our model predicts future 6-DOF object trajectories conditioned on a specified goal state.
    We encode (a) trajectory dynamics, (b) local geometry propagated from the scene point cloud to the object's bounding box, (c) semantic fixture boxes and labels, (d) natural language description of the action (e) a goal descriptor.
    A multimodal transformer performs hierarchical fusion that emphasizes geometric feasibility before semantic preferences.
    The fused latent is fed directly to the prediction head (no separate decoding stage), which we found more stable for long-horizon control.
    }
    \label{fig:pipeline}
    \vspace{-1em}
\end{figure*}

Our goal is to synthesize controllable 6-DOF object trajectories in 3D scenes, conditioned on observed motion, scene context, and a target goal state. The central challenge is to fuse heterogeneous modalities: geometry, semantics, and dynamics into a single representation that preserves physical plausibility and goal consistency. Naively concatenating features or relying on a single modality (e.g., raw point clouds) caused unstable training and implausible motion (e.g., interpenetration, drifting). We design a multimodal transformer with three key insights: (1) \emph{spatial feature propagation} is a compact yet stable spatial abstraction compared to dense point features; (2) \emph{vision language semantics} (CLIP) transfer behavior patterns across action description or categories better than one-hot labels; and (3) \emph{hierarchical fusion} that prioritizes hard geometric constraints over softer semantic cues significantly reduces collision and goal drift. An overview is shown in Fig.~\ref{fig:pipeline}.

\subsection{Problem Formulation}
\label{subsec:problem}
We formulate trajectory prediction as a conditional sequence modeling problem. 

Let the input history trajectory be denoted as 
$\mathbf{X}_{1:H}=\{\mathbf{x}_1, \mathbf{x}_2, \ldots, \mathbf{x}_H\} \in \mathbb{R}^{H \times 9}$, where each $\mathbf{x}_i = (\mathbf{p}_i, \mathbf{r}_i)$ consists of a 3D position $\mathbf{p}_i \in \mathbb{R}^3$ and a 6D continuous rotation representation $\mathbf{r}_i \in \mathbb{R}^6$~\cite{zhou2019continuity}. 
Each object trajectory is associated with a fixed object category $c$ and size $s \in \mathbb{R}^3$.

The scene context is represented as $\mathbf{S} = (\mathbf{P}, \mathbf{B})$, where $\mathbf{P} \in \mathbb{R}^{N \times 3}$ denotes the scene point cloud with $N$ points, $\mathbf{B} = \{ (l_k, b_k) \}_{k=1}^{M}$ is defined as the set of $M$ semantic fixture bounding boxes,  
with each $l_k$ being the semantic label, and $b_k = (\mathbf{c}_k, \mathbf{s}_k, \mathbf{r}_k) \in \mathbb{R}^{12}$ containing the 3D center, size, and 6D rotation representation.
The goal condition $\mathbf{G} \in \mathbb{R}^9$ specifies the target object state.
The output trajectory to be predicted is $\hat{\mathbf{X}}_{H+1:T} = \{\hat{\mathbf{x}}_{H+1}, \hat{\mathbf{x}}_{H+2}, \ldots, \hat{\mathbf{x}}_{H}\}$, where $T-H$ is the prediction horizon. In our particular setup, we use 30\% of the trajectory as the input history to predict the remaining 70\%. 
Thus, the task is to learn a conditional distribution $P(\hat{\mathbf{X}}_{H+1:T} \mid \mathbf{X}_{1:H}, \mathbf{G}, \mathbf{S})$, which generates physically plausible and semantically consistent future trajectories, aligned with the specified goal and conditioned on multimodal scene context.

\subsection{Multimodal Scene Encoding}
\label{subsec:encoding}
Generating plausible object trajectories requires a comprehensive understanding of both spatial and semantic context. Specifically, the model must capture (1) object motion patterns, (2) spatial arrangements, and (3) environmental constraints, including collision and interaction dynamics. To this end, we construct dedicated representations for each scene modality.

\noindent\textbf{Trajectory Feature.}
We observe that using trajectory geometry alone often leads to overfitting, as the model fails to capture category-specific motion patterns. To address this, we couple trajectory embeddings with semantic category features.

The temporal motion context $\mathbf{F}_t$ is obtained by passing the observed trajectory sequence through a linear layer, yielding an embedding suitable for multimodal fusion. 

\noindent\textbf{Spatial Feature.}
To account for environmental constraints such as floors, walls, or tabletops, the model must capture both global and local spatial cues. Directly concatenating a global scene feature $\mathbf{F}_o$ is insufficient, as it fails to distinguish spatially distinct regions (e.g., floor vs. tabletop) and introduce irrelevant noise (e.g., clutter on the ground). Instead, we encode the raw point cloud $\mathbf{P}$ using PointNet++~\cite{qi2017pointnet++}, producing both a global feature $\mathbf{F}_o$ and local features $\mathbf{F}_l$.
To provide trajectory features with awareness of their local surroundings, we propagate per-point local features from the scene point cloud to the object's bounding box at each observed timestep. Specifically, we interpolate features from the $k$ nearest neighbors using inverse-distance weighting~\cite{qi2017pointnet++}:
\begin{equation}
\mathbf{F}^t_{p} = \frac{\sum_{i=1}^k w_i(\mathbf{c}_t)\mathbf{f}_i}{\sum_{i=1}^k w_i(\mathbf{c}_t)}, \quad
w_i(\mathbf{c}_t) = \frac{1}{|\mathbf{c}_t - \mathbf{p}_i|^2},
\end{equation}
where $\mathbf{c}_t$ denotes the center of the object bounding box at time $t$, and $\mathbf{p}_i$ and $\mathbf{f}_i$ are the coordinates and features of the $i$-th point, respectively.
By repeating this process over all observed timesteps $t = 1, \ldots, H$, we obtain a temporal sequence of local geometric features $\mathbf{F}_{p} = \{ \mathbf{F}_{p}^1, \mathbf{F}_{p}^2, \ldots, \mathbf{F}_{p}^{H} \}$, which are subsequently incorporated in the fusion stage to enrich trajectory representations with spatial context.

Furthermore, relying solely on point-cloud features is insufficient to fully capture the interaction constraints between the moving object and nearby static fixtures. The spatial extents of these fixtures can be reliably obtained using modern instance segmentation approaches~\cite{ren2024grounded, khanam2024yolov11, carion2025sam}. Direct concatenation of their embeddings, however, fails to model inter-object relationships and may result in physically implausible predictions (e.g., a chair penetrating a tabletop). To more explicitly encode such interaction constraints, we apply a multi-head self-attention module over the set of fixture bounding boxes:
\begin{equation}
    \mathbf{F}_b = \mathrm{SelfAttn}\big(\{b_k\}_{k=1}^M\big).
\end{equation}

\noindent\textbf{Semantic Feature.}
Geometry alone is insufficient to distinguish objects with similar sizes but different affordances (e.g., desk vs. bed). Semantic information is incorporated by embedding fixture labels $l_k$ and the natural language description of the action/object category name $d$ involving the object with a frozen CLIP encoder~\cite{radford2021learning}, followed by projection into the feature space:
\begin{equation}
\mathbf{F}_f= \mathrm{Proj}(\mathrm{CLIP}(l_k)), \mathbf{F}_d= \mathrm{Proj}(\mathrm{CLIP}(d))
\end{equation}
To mitigate semantic noise, only the $K$ nearest fixtures (based on center distance) are retained before applying attention, as distant objects contribute little and increase variance.

\noindent\textbf{Goal Feature.}
The target object state $\mathbf{G}$, representing the desired future position and orientation, is encoded via a linear layer and projected into the same feature space:
\begin{equation}
\mathbf{F}_g = \mathrm{Proj}(\mathrm{Linear}(\mathbf{G})).
\end{equation}
Note that this goal plays two complementary roles.
First, it serves as a high-level intention variable that disambiguates between otherwise plausible futures under the same history and scene, e.g., ``place the object on the tabletop'' versus ``place it back on the floor''.
Second, it provides a controllable knob at inference time: given a fixed observed trajectory $\mathbf{X}_{1:H}$ and scene $\mathbf{S}$, different choices of $\mathbf{G}$ induce qualitatively different yet physically valid futures.
By making $\mathbf{G}$ a first-class conditioning signal, our formulation bridges trajectory prediction and goal-directed planning, enabling the synthesis of future motions that are not only plausible and scene-consistent but also explicitly aligned with user-specified targets.

\subsection{Multimodal Feature Fusion}
\label{subsec:fusion}

\noindent\textbf{Multi-modal Transformer.}
Naively concatenating features across modalities can cause scale imbalance, leading the model to over-rely on certain inputs. To achieve balanced and flexible integration, we adopt a Transformer-based fusion module inspired by Perceiver IO~\cite{jaegle2021perceiver}. This design introduces a learnable latent array that acts as an information bottleneck, ensuring scale normalization across modalities and enabling modality-agnostic fusion.

Given a collection of modality-specific input tokens $\mathbf{X} = \{ \mathbf{x}_1, \ldots, \mathbf{x}_M \}$, where $\mathbf{X} \in \mathbb{R}^{M \times d_{\text{in}}}$, the fusion module maintains a learnable latent array $\mathbf{Z}_0 \in \mathbb{R}^{N \times d_{\text{latent}}}$, with $N \ll M$ and $d_{\text{latent}}$ denoting the latent feature dimension.

The fusion process is implemented via stacked cross-attention and latent self-attention blocks. At each layer $\ell$, the latent array is first updated by attending to the input tokens:
\begin{equation}
\label{eq:crossattn}
\mathbf{Z}_{\ell}' = \mathrm{CrossAttn}(\mathbf{Z}_{\ell-1}, \mathbf{X}) = \mathrm{softmax}\left( \frac{QK^\top}{\sqrt{d_K}} \right) V,
\end{equation}
where $Q = \mathbf{Z}_{\ell-1} W_q$, $K = \mathbf{X} W_k$, and $V = \mathbf{X} W_v$ are linearly projected query, key, and value matrices, respectively.

Next, latent self-attention and a feed-forward network are applied to propagate information among latent slots:
\begin{equation}
\label{eq:selfattn}
\mathbf{Z}_{\ell} = \mathrm{FFN}\left( \mathrm{SelfAttn}(\mathbf{Z}_{\ell}') \right),
\end{equation}
where $\mathrm{SelfAttn}$ follows the same formulation as Eq.~\ref{eq:crossattn}, but operates only on latent tokens.

After $L$ layers, the final latent representation $\mathbf{Z}_L$ serves as the fused multimodal embedding. Unlike Perceiver IO~\cite{jaegle2021perceiver}, our architecture does not include a second decoding stage; instead, $\mathbf{Z}_L$ is directly used as input to the trajectory prediction head.

\noindent\textbf{Scene-aware Trajectory Augmentation.}
To endow trajectory features with spatial awareness, we fuse $\mathbf{F}_t$ with the propagated local geometry $\mathbf{F}_{p}^t$ through the multimodal transformer:
\begin{equation}
\mathbf{F}_{tp}=\mathrm{MultiTrans}\big(\mathrm{Concat}(\mathbf{F}_t,\mathbf{F}_{p}^t)\big).
\end{equation}
Direct concatenation without attention underperformed, indicating that attention-based alignment is necessary to resolve frame and scale ambiguities.

\noindent\textbf{Semantic Geometric Fusion.} While the trajectory features are enhanced with 3D spatial information, incorporating semantic cues is crucial for guiding trajectory generation, as the prediction network must understand object semantics and their relations (e.g., a chair should not move through a table). To this end, we project the semantic features $\mathbf{F}_f$ and $\mathbf{F}_d$, the spatial features $\mathbf{F}_b$, the global point cloud feature $\mathbf{F}_o$, and the goal state feature $\mathbf{F}_g$ into the same dimension as the fused trajectory feature $\mathbf{F}_{tp}$ via linear layers. These features are then concatenated to form a comprehensive multimodal representation:
\begin{equation}
    \mathbf{F}_{\text{fuse}} = \mathrm{Concat}(\mathbf{F}_{tp}, \mathbf{F}_f, \mathbf{F}_d, \mathbf{F}_b, \mathbf{F}_o, \mathbf{F}_g).
\end{equation}
Finally, the predicted trajectory $\hat{\mathbf{X}}_{H:T}$ is conditioned on the fused feature $\mathbf{F}_{\text{fuse}}$ through the multimodal transformer described above.

\subsection{Training Objective}

Our training objective is designed to supervise both the future prediction accuracy and the reconstruction quality of the observed motion. The total loss comprises four terms: translation loss, orientation loss, reconstruction loss, and destination loss for both translation and orientation. Each term captures a specific aspect of the prediction quality.

Given the model's prediction $\hat{\mathbf{X}} \in \mathbb{R}^{T \times 9}$ and the ground truth trajectory $\mathbf{X} \in \mathbb{R}^{T \times 9}$, where each frame contains 3D position and 6D rotation representation, we first split each sequence into history and future segments based on predefined input and predict ratios. Let $T_\text{hist}$ and $T_\text{fut}$ denote the number of historical and future steps, respectively.

To supervise the \textit{future prediction}, we compute an $L_1$ loss between the predicted and ground truth values for both translation and rotation components:
\begin{align}
\mathcal{L}_\text{trans} &= \frac{1}{T_\text{fut}} \sum_{t \in \text{future}} \left\| \hat{\mathbf{p}}_t - \mathbf{p}_t \right\|_1, \\
\mathcal{L}_\text{ori} &= \frac{1}{T_\text{fut}} \sum_{t \in \text{future}} \left\| \hat{\mathbf{r}}_t - \mathbf{r}_t \right\|_1,
\end{align}
where $\mathbf{p}_t \in \mathbb{R}^3$ and $\mathbf{r}_t \in \mathbb{R}^6$ denote the ground truth position and rotation at timestep $t$, and $\hat{\mathbf{p}}_t$, $\hat{\mathbf{r}}_t$ are their corresponding predictions.

To preserve fidelity in the \textit{observed segment}, a reconstruction loss is applied to the historical frames:
\begin{equation}
\mathcal{L}_\text{rec} = \frac{1}{T_\text{hist}} \sum_{t \in \text{history}} \left\| \hat{\mathbf{x}}_t - \mathbf{x}_t \right\|_1,
\end{equation}
where $\hat{\mathbf{x}}_t$ is the full 9D predicted pose at timestep $t$.

Additionally, we introduce a \textit{destination loss} to explicitly constrain the model's final predicted pose to match the last valid ground truth frame:
\begin{equation}
\mathcal{L}_\text{dest} = \left\| \hat{\mathbf{x}}_{T_\text{end}} - \mathbf{x}_{T_\text{end}} \right\|_1,
\end{equation}

The final loss is a weighted sum of the above terms:
\begin{equation}
\mathcal{L}_\text{total} = \lambda_\text{trans} \mathcal{L}_\text{trans} + \lambda_\text{ori} \mathcal{L}_\text{ori} + \lambda_\text{rec} \mathcal{L}_\text{rec} + \lambda_\text{dest} \mathcal{L}_\text{dest},
\end{equation}
where $\lambda_\text{trans}, \lambda_\text{ori}, \lambda_\text{rec}, \lambda_\text{dest}$ are hyperparameters controlling the contribution of each loss component.

%% file: sec/4_experiment.tex
\section{Experiments}
\label{sec:exp}

\begin{figure*}[t]
    \centering
    \includegraphics[width=1\textwidth]{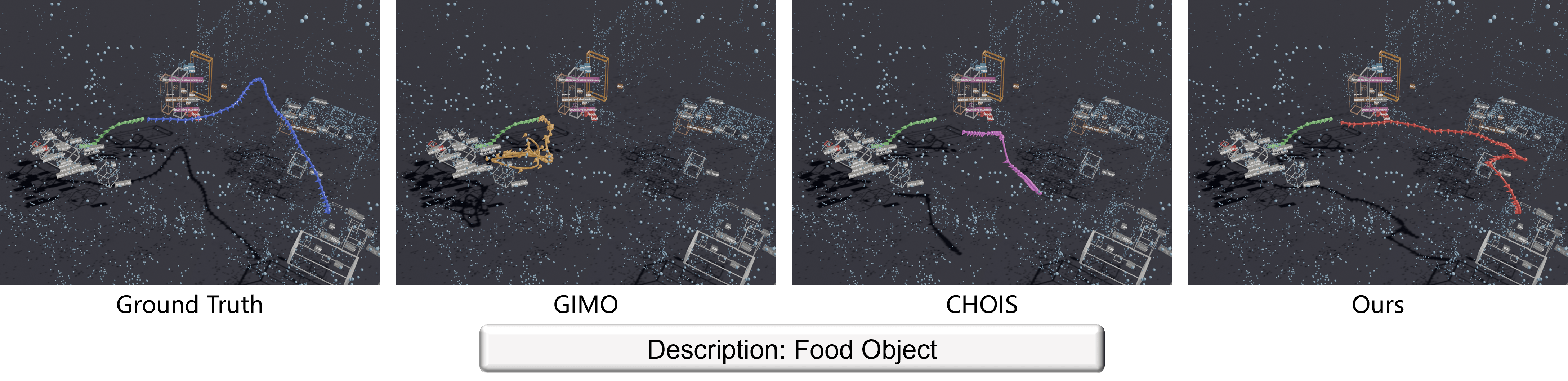}
    \caption{
        \textbf{Qualitative results on the ADT dataset.} 
        The green trajectory represents the input history across all experiments. Only our model produces trajectories that both reach the target and avoid collisions, while also achieving shorter path lengths compared to the ground-truth natural trajectories. Adaptive GIMO fails due to the absence of gaze information, whereas CHOIS accumulates errors over time, ultimately leading to failure.
    }
    \label{fig:qualitative_adt}
\end{figure*}

We start this section by describing the experimental setup, including datasets and metrics. We then present results on two datasets, followed by an ablation study to analyze the contributions of different components in our method.
\subsection{Experiments Setup}
\label{subsec:setup}

\noindent\textbf{Baseline.} 
To provide comparative insights, we adapt two state-of-the-art human motion prediction methods for the object manipulation trajectory generation task:

\begin{itemize}
    \item \textbf{GIMO}~\cite{zheng2022gimo}: A transformer-based model originally designed for egocentric human-object interaction forecasting. It leverages a unified Perceiver-inspired architecture to fuse geometry, object category, and semantic scene context for predicting 6-DOF human motion. We re-purpose GIMO to predict object trajectories by replacing the human body input with object-specific motion and geometry. Since our object-centric prediction task does not contain gaze information, we disable the gaze branch and remove all gaze-related modules in both training and inference.
    \item \textbf{CHOIS}~\cite{li2024controllable}: This generative framework produces human-object interaction sequences conditioned on object geometry, sparse object waypoints, and textual instructions. In our implementation, the human-motion branch is deactivated, preserving only the object trajectory prediction component. The waypoint conditioning is restricted to the initial 30\% of the input and goal for fair comparison. To ensure compatibility with our deterministic prediction framework, we disable the diffusion-based sampling mechanism and utilize only the transformer backbone for direct single-step prediction.
\end{itemize}

\vspace{0.5em}

\noindent\textbf{Metrics.}
We evaluate predicted object trajectories using the following quantitative metrics to assess accuracy, temporal consistency, and physical plausibility:

\begin{itemize}
    \item \textbf{Average Displacement Error} (ADE): Measures the mean L2 distance between the predicted and ground truth object positions across all future time steps.

    \item \textbf{Final Displacement Error} (FDE): L2 distance between the predicted and ground truth position at the final future time step.

    \item \textbf{Fréchet Distance}~\cite{efrat2002new} (FD): Measures the maximum deviation between two trajectories over time by considering the best possible alignment along the temporal axis. It is sensitive to both spatial proximity and temporal consistency. A smaller Fréchet distance indicates that the predicted trajectory closely follows the shape and timing of the ground truth, whereas a large value indicates temporal mismatch or outlier deviations.

    \item \textbf{Angular Consistency} (AC): Measures how well the directional dynamics of the predicted trajectory align with the reference sequence. The positional differences between consecutive frames are represented as direction vectors, and the mean cosine similarity between these vectors quantifies the preservation of orientation trends and motion smoothness. Higher values indicate better temporal coherence and directional stability. 

    \item \textbf{Collision Rate} (CR): The fraction of predicted trajectories that result in collisions with surrounding fixtures, computed based on the intersection of predicted bounding boxes and static scene geometry. Lower collision rates indicate better physical plausibility and spatial awareness of the model.
\end{itemize}

\subsection{Controlled Idealized Scenarios}
\label{subsec:adt}
\input{tables/adt}
\noindent\textbf{Dataset.}   
We first evaluate our method on the Aria Digital Twin (ADT) dataset~\cite{pan2023aria}, which offers high-fidelity recordings of human-object interactions in a fully controlled 3D simulation environment. The sequences are captured under noise-free conditions with complete visibility and accurate tracking, providing an ideal setting to assess the upper-bound performance of our model under perfect geometric and semantic observations. To align with our task objective, we exclude all ADT sequences involving recognition-only interactions (i.e., without significant object displacement). We train our model and the baselines on a randomly selected subset of 228 sequences and evaluate them on the remainder.

\noindent\textbf{Results.}
As shown in Tab.~\ref{tab:adt}, our method consistently outperforms both adapted baselines across all evaluation metrics except for the collision rate, with significant gains in spatial accuracy (ADE/FDE). Notably, it achieves the lowest Fréchet distance and highest angular consistency, indicating superior alignment with the ground-truth in both position and orientation. While a lower collision rate can be trivially achieved by predicting static trajectories, only our model attains both low FDE and a low collision rate, demonstrating its ability to generate plausible and physically meaningful object motions. An illustration is shown in Fig.~\ref{fig:qualitative_adt} and more in the supplementary.

\begin{figure*}[t]
    \centering
    \includegraphics[width=1\textwidth]{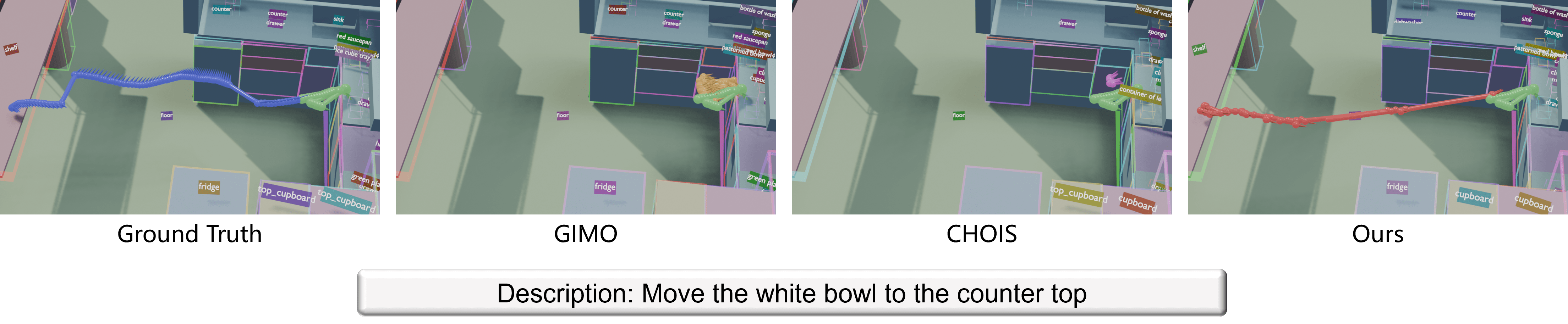}
    \caption{
    \textbf{Qualitative results on the HD-EPIC dataset.} 
    Across all examples, the green points indicate the input history. Our model generates trajectories that are more efficient than the ground truth, while all baselines remain stuck in repetitive motions.
    }
    \label{fig:qualitative_hdepic}
\end{figure*}

\subsection{Realistic Challenging Scenarios}
\label{subsec:hdepic}
\input{tables/hdepic}
\noindent\textbf{Dataset.}
To assess the robustness of our model in real-world settings, we evaluate it on the HD-EPIC dataset~\cite{perrett2025hd}. HD-EPIC contains $41$ hours of egocentric videos of human-object interactions recorded in natural household environments. 
A key challenge in leveraging the HD-EPIC dataset for our task is the sparsity of its annotations. The dataset provides object positions for pickup and drop events, but does not provide dense, frame-by-frame object trajectory sequences between these points. To bridge this gap, we use the interacting hand as a proxy for computing the object's motion. Our key assumption is that between the moments of pickup and release, the hand and object are physically coupled and move together. By tracking the hand's motion using Project Aria's Machine Perception Service (MPS)~\cite{engel2023project}, we can accurately infer the object's trajectory during this unannotated phase.

Our process begins by identifying the primary hand interacting with the object. At the start of each sequence, we use the dataset's initial 3D object position to select the hand with the minimum Euclidean distance to the annotated object position. For all subsequent frames until the drop-off, this choice is propagated by identifying which hand is in physical contact, as determined by using a pretrained Hands23~\cite{cheng2023hands23} detector. A sliding window filter is then applied to this sequence of hand selections to ensure better temporal consistency and remove flickering.

Finally, we process the motion of this consistently tracked hand to create a stable trajectory. We compute a 6D rotation for the hand's orientation by constructing an orthonormal basis from its normal vector of the wrist-palm using singular value decomposition (SVD). This refined 6-DOF hand trajectory is then directly transferred to the annotated target object, yielding a clean and realistic manipulation sequence.

\noindent\textbf{Results.}
Tab.~\ref{tab:hdepic} summarizes the quantitative results. Our model consistently and significantly outperforms all baselines across all metrics. Although the HD-EPIC dataset poses greater challenges due to occlusions and sensor noise, its substantially larger scale ($\sim$20$\times$ trajectories than ADT) and more diverse scenes help compensate for these difficulties and allow the model to maintain strong performance. Examples can be found in Fig.~\ref{fig:qualitative_hdepic} and in the supplementary.

\subsection{Ablation Study}
\label{subsec:ablation}
\input{tables/ablation}
To analyze the contributions of different components in our method, we conduct an ablation study on the ADT dataset. We first evaluate the effect of each input modality by removing geometric and semantic information individually. Next, we examine the influence of input trajectory length by reducing the number of observed frames to using only the first frame. Finally, we assess the importance of goal conditioning by removing the goal input from the model.

\noindent\textbf{Results.}
Table~\ref{tab:ablation} reports the ablation results on the ADT dataset. Removing geometric features leads to a noticeable degradation in ADE and Fréchet distance, indicating that local spatial structure is important for accurate trajectory prediction. Excluding semantic information similarly worsens overall performance, though the model retains reasonable final-position accuracy. In contrast, removing goal conditioning results in a substantial drop across all metrics, confirming that explicit goal specification is essential for producing coherent long-horizon trajectories. The ``first frame'' variant performs the worst, with extremely high collision rates and large deviations from the target, demonstrating that dynamic context is crucial for stable motion generation. Overall, the full model achieves the best Fréchet distance, angular similarity, and lowest collision rate, highlighting the complementary contributions of geometric, semantic, and goal-related features.

%% file: tables/adt.tex
\begin{table}[h]
    \centering
    \scriptsize
    \begin{tabular}{lccccc}
        \toprule
        Method & ADE[m] ↓ & FDE[m] ↓ & FD[m] ↓ & AC[m] ↑ & CR ↓ \\
        \midrule
        GIMO~\cite{zheng2022gimo}  & 0.982 & 1.401 & 1.511 & 0.140 & 19.6\% \\
        CHOIS~\cite{li2024controllable} & 0.853 & 1.062 & 1.209 & 0.283 & \textbf{9.3\%} \\
        \methodname(Ours)   & \textbf{0.366} & \textbf{0.072} & \textbf{0.438} & \textbf{0.402} & 13.1\% \\
        \bottomrule
    \end{tabular}
    \caption{Quantitative results on the ADT dataset. Our method outperforms both baselines in all metrics except collision rate. Note that collision rate is only meaningful when FDE is also low; otherwise, it may decrease due to trivial predictions such as static forecasts.}
    \label{tab:adt}
\end{table}

%% file: tables/hdepic.tex
\begin{table}[h]
    \centering
    \scriptsize
    \begin{tabular}{lccccc}
        \toprule
        Method & ADE[m] ↓ & FDE[m] ↓ & FD[m] ↓ & AC[m] ↑ & CR ↓ \\
        \midrule
        GIMO~\cite{zheng2022gimo}  & 0.411 & 0.654 & 0.780 & 0.002 & 11.8\% \\
        CHOIS~\cite{li2024controllable}  & 0.446 & 0.589 & 0.760 & 0.009 & 12.0\% \\
        \methodname(Ours)   & \textbf{0.283} & \textbf{0.034} & \textbf{0.391} & \textbf{0.037} & \textbf{10.3\%} \\
        \bottomrule
    \end{tabular}
    \caption{Quantitative results on the HD-EPIC dataset. Ours achieves the best performance across all metrics, demonstrating superior trajectory prediction accuracy and robustness in real-world scenarios.}
    \label{tab:hdepic}
\end{table}

%% file: tables/ablation.tex
\begin{table}[h]
    \centering
    \scriptsize
    \begin{tabular}{lccccc}
        \toprule
        Variant & ADE[m] ↓ & FDE[m] ↓ & FD[m] ↓ & AC[m] ↑ & CR ↓ \\
        \midrule
        w/o pointcloud      & 0.364 & \textbf{0.062} & 0.493 & 0.384 & 18.7\% \\
        w/o semantic  & \textbf{0.360} & 0.080 & 0.505 & 0.391 & 14.0\% \\
        w/o goal         & 0.531 & 0.593 & 0.729 & 0.311 & 13.3\% \\
        First frame  & 0.554 & 0.258 & 0.696 & 0.298 & 83.2\% \\
        \methodname(Full)   & 0.366 & 0.072 & \textbf{0.438} & \textbf{0.402} & \textbf{13.1\%} \\
        \bottomrule
    \end{tabular}
    \caption{Ablation study on the ADT dataset. The best results are highlighted in bold.}
    \label{tab:ablation}
\end{table}

%% file: sec/5_conclusion.tex
\section{Limitations, Future Work and Conclusion}
In this work, we introduced a trajectory-centric framework that predicts realistic and controllable 6-DOF object trajectories in complex 3D environments. By combining geometric representations, semantic cues, and goal conditioning, our model bridges perception and control through a flexible and generalizable formulation. A key insight of our approach is that object trajectories themselves serve as an effective intermediate representation, enabling cross-embodiment execution via inverse kinematics and simplifying the integration of downstream planners. This design allows us to achieve accurate spatial reasoning and efficient trajectory synthesis while maintaining broad applicability.

Despite these advances, our work has several limitations. First, the model assumes well-aligned scene context and object annotations, which may not hold in cluttered or noisy real-world scenarios.
Second, the provided goal condition plays a decisive role in guiding the trajectory generation process; however, such information is often unavailable or difficult to obtain in real applications.

Future work will address these limitations by incorporating goal inference mechanisms that estimate plausible target states from visual observations and contextual cues (e.g. VLM). 
Furthermore, integrating reinforcement learning or closed-loop feedback could improve adaptation to unseen conditions and support long-horizon planning. Additionally, introducing post refinement based on collision optimization is also a feasible direction. We also see potential in exploring broader cross-embodiment transfer, where a single predicted object trajectory can guide manipulation across robots with different morphologies.

\noindent\textbf{Acknowledgments.} This research was partially funded by the German Federal Ministry of Education and Research through the ExperTeam4KI funding program for UDance (Grant No. 16IS24064).

%% file: sec/X_suppl.tex
\clearpage
\setcounter{page}{1}
\maketitlesupplementary

In this supplementary material, we provide detailed descriptions of our implementation, dataset processing procedures, and additional experimental results. The document is organized as follows. We first outline the implementation details in Sec.~\ref{sec:supp_implementation}, then the overall dataset processing workflow in Sec.~\ref{sec:supp_dataset_processing}, followed by the dataset-specific pre-processing steps applied to the Aria Digital Twin (ADT)~\cite{pan2023aria} and HD-EPIC~\cite{perrett2025hd} datasets. Next, in Sec.~\ref{sec:supp_qualitative_results}, we present several failure cases arising from complex motion patterns, along with additional qualitative results.

\section{Implementation Details}
\label{sec:supp_implementation}

Our model is implemented in PyTorch~\cite{paszke2019pytorch} and is trained on a single NVIDIA A6000 GPU. The training process utilizes the AdamW optimizer with a learning rate of $1 \times 10^{-4}$, weight decay of $5 \times 10^{-4}$, and exponential learning rate decay with a factor of $0.99$.

For data usage, $90\%$ of the available sequences are used for training, with the remaining $10\%$ split equally between validation and testing. 
Each training sample consists of a full scene point cloud, semantic fixture bounding boxes and labels, and a 6-DOF object trajectory. 
Trajectories are uniformly resampled to $200$ frames, and the first $30\%$ of the sequence is used as the input history.

The multimodal Transformer encoder consists of $6$ layers with $8$ attention heads per layer. 
The input trajectory is embedded in a $128$-dimensional feature space; the global scene point cloud feature has $128$ dimensions, and per-point features have $64$ dimensions. Semantic fixture bounding boxes are projected to $128$ dimensions, while CLIP~\cite{radford2021learning} embeddings for object categories and semantic labels are linearly projected to the same dimension. 
The goal state feature is also projected to $128$ dimensions to ensure compatibility in the fusion space. 
The latent array within the Transformer is $256$-dimensional.

During training, the loss function combines translation, orientation, reconstruction, and destination losses, with weights $\lambda_\text{trans}$, $\lambda_\text{ori}$, $\lambda_\text{rec}$, and $\lambda_\text{dest}$ set to 1.0. 

\section{Dataset Processing}
\label{sec:supp_dataset_processing}

\noindent\textbf{General Processing.}
We begin by processing the trajectory data. Since trajectories in ADT are overly dense, we downsample them by retaining one point every five frames to improve computational efficiency while preserving essential motion cues. To support training with multiple batches, we fix the predicted trajectory length to 200. Trajectories shorter than 200 frames are padded, whereas longer ones are truncated. An attention mask ensures that only valid trajectory points contribute to the training objective. We further apply two filtering rules to discard unrealistic motion: an object is considered to be moving only when its velocity exceeds 0.05\,m/s, and segments are labeled static when an object remains still for more than three consecutive frames. These criteria help preserve only meaningful trajectories for both training and evaluation.

Next, instead of using the full scene point cloud, we extract only the local region around each trajectory to reduce computational overhead. For every trajectory point, we define a spherical neighborhood with a radius of 1\,m and collect all points within this region, ensuring that only relevant scene geometry is retained. Similarly, fixture information is extracted only from regions near the trajectory, as distant geometry has limited influence on downstream tasks. These extracted fixture features are then incorporated into the model input. We now describe dataset-specific processing procedures for ADT and HD-EPIC.

\noindent\textbf{Aria Digital Twin Dataset.}
Since ADT does not include action-level semantic annotations, we use only object categories as descriptive information. Certain trajectories are removed because they do not meet the requirements of our robotic manipulation setting. For example, trajectories in which an object is held or moved within an extremely limited spatial region for extended periods are excluded, as they lack meaningful interaction patterns and do not reflect practical robotic behaviors.
\begin{figure*}[t]
    \centering
    \includegraphics[width=1\textwidth]{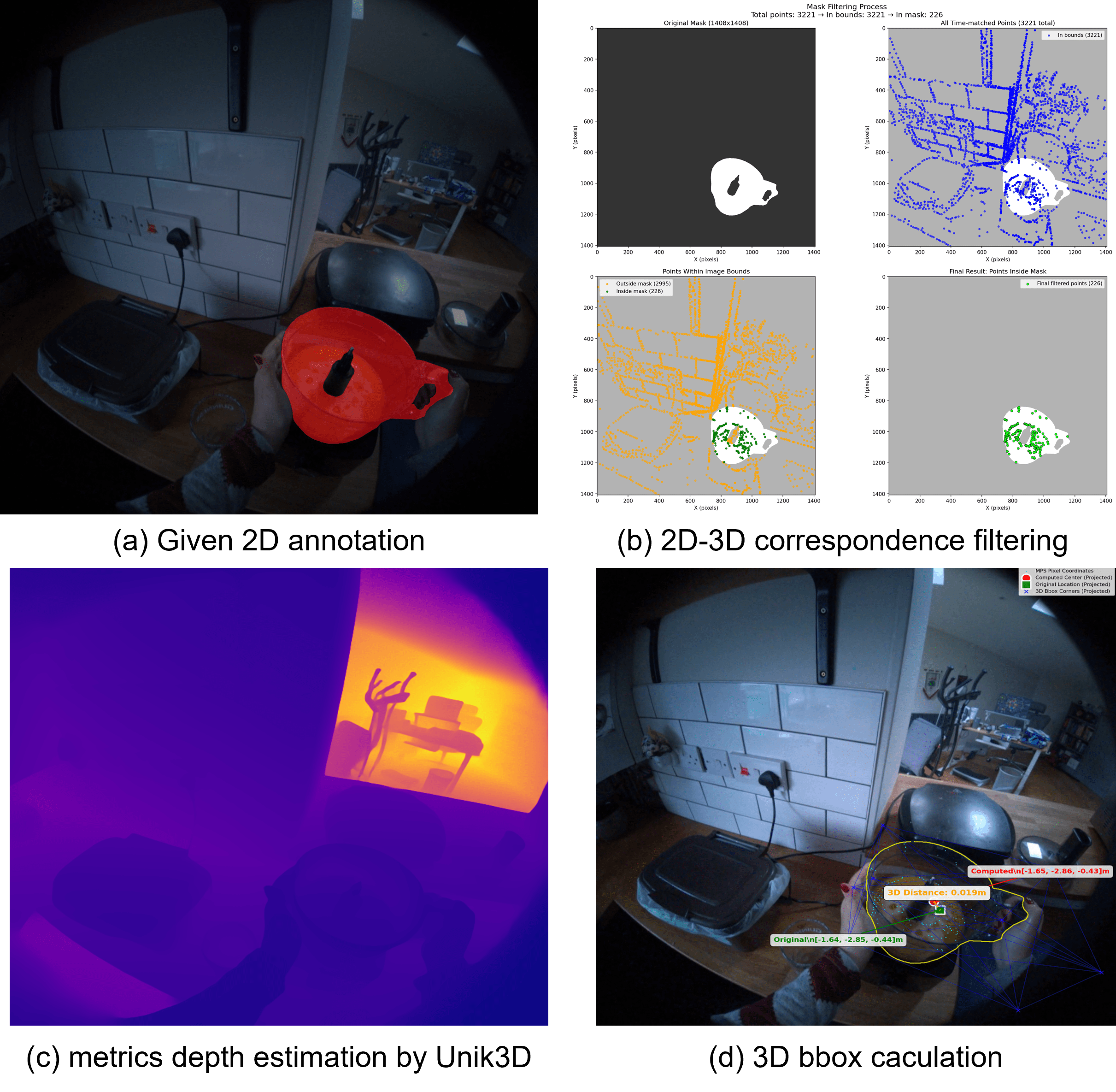}
    \caption{
    \textbf{3D bounding box reconstruction in the HD-EPIC dataset.} 
    (a): input RGB frame with object mask. 
    (b): mask filtering and sparse 2D-3D correspondences from SLAM and MPS data. 
    (c): monocular depth estimation from UniK3D. 
    (d): final 3D bounding box recovered after depth alignment and scaling. 
    This pipeline enables accurate localization of small objects (e.g., bowls, cups) in cluttered scenes.
    }
    \label{fig:bbox_reconstruction}
\end{figure*}

\noindent\textbf{HD-EPIC Dataset.}
In contrast to ADT, the HD-EPIC dataset does not provide bounding boxes or semantic labels for scene objects. Large static structures such as countertops and drawers are manually reconstructed in Blender and aligned with the scene point cloud to serve as static bounding boxes. For small manipulable objects such as coffee machines and knives, HD-EPIC provides start/end timestamps of object motion, 2D masks, and 3D object centers. Using this information, we first align the timestamps with SLAM data and obtain sparse 2D--3D correspondences using MPS data collected with Aria glasses. We then estimate monocular depth using UniK3D~\cite{piccinelli2025unik3d}, perform linear depth alignment with the correspondences to recover the true scale, and reconstruct 3D object bounding boxes. These \textit{dynamic bounding boxes}, together with static ones, are used for model training. Fig.~\ref{fig:bbox_reconstruction} illustrates the full processing pipeline.

Since the dataset provides only pick-up and drop-off annotations for each object, we generate dense object trajectories using hand-tracking. We compute the transformations of the manipulating hand over time and apply these transformations to the provided initial object position, producing a complete trajectory for each object.

The hand trajectory extraction pipeline integrates multiple perception systems to process egocentric video. The pipeline begins with a bootstrap stage that establishes hand--object correspondence by computing the 3D Euclidean distance between detected hand positions and the annotated object center in world coordinates to identify the primary manipulating hand. Subsequent hand positions are obtained using Project Aria's MPS~\cite{engel2023project}. For frames in which both hands are confidently detected, we leverage Hands23~\cite{cheng2023hands23} to disambiguate which hand is physically interacting with the object. Hands23 infers binary contact states for each hand, enabling reliable determination of the manipulating hand even when both hands appear in view. When Hands23 outputs are ambiguous (e.g., both hands detected in contact), the system maintains temporal consistency by defaulting to the initially selected primary hand. Temporal coherence is further enforced via a sliding-window filter (window size = 3), which suppresses spurious frame-to-frame switching.

For orientation estimation, we construct a 6D rotation representation~\cite{zhou2019continuity} derived from the geometric structure of the hand. Specifically, we use Singular Value Decomposition (SVD) to compute the hand coordinate frame, where the primary axis aligns with the wrist-to-palm vector and the palm normal defines the facing direction, both obtained from MPS. This 6D parameterization ensures continuity across the rotation manifold and avoids the singularities present in Euler angles and quaternions.

\begin{figure*}[t]
    \centering
    \includegraphics[width=1\textwidth]{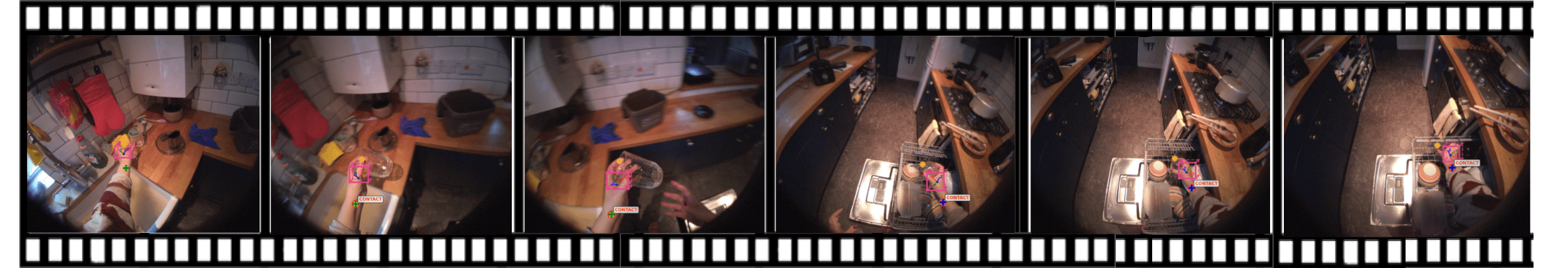}
    \caption{
    \textbf{Object Position computation using hand-tracking} 
    We demonstrate the object positions. depicted as Yellow \* along with the orientation and the hand that is interacting with the object at the particular frame sampled at intervals during the entire period of the moving object. 
    }
    \label{fig:hdepic_hands}
\end{figure*}

We demonstrate the effectiveness of the reconstructed object trajectories in Fig.~\ref{fig:hdepic_hands}, where the recovered 3D object location is projected onto the input RGB frames for visual verification.

\section{Additional Qualitative Results}
\label{sec:supp_qualitative_results}
\begin{figure*}[t]
    \centering
    \includegraphics[width=1\textwidth]{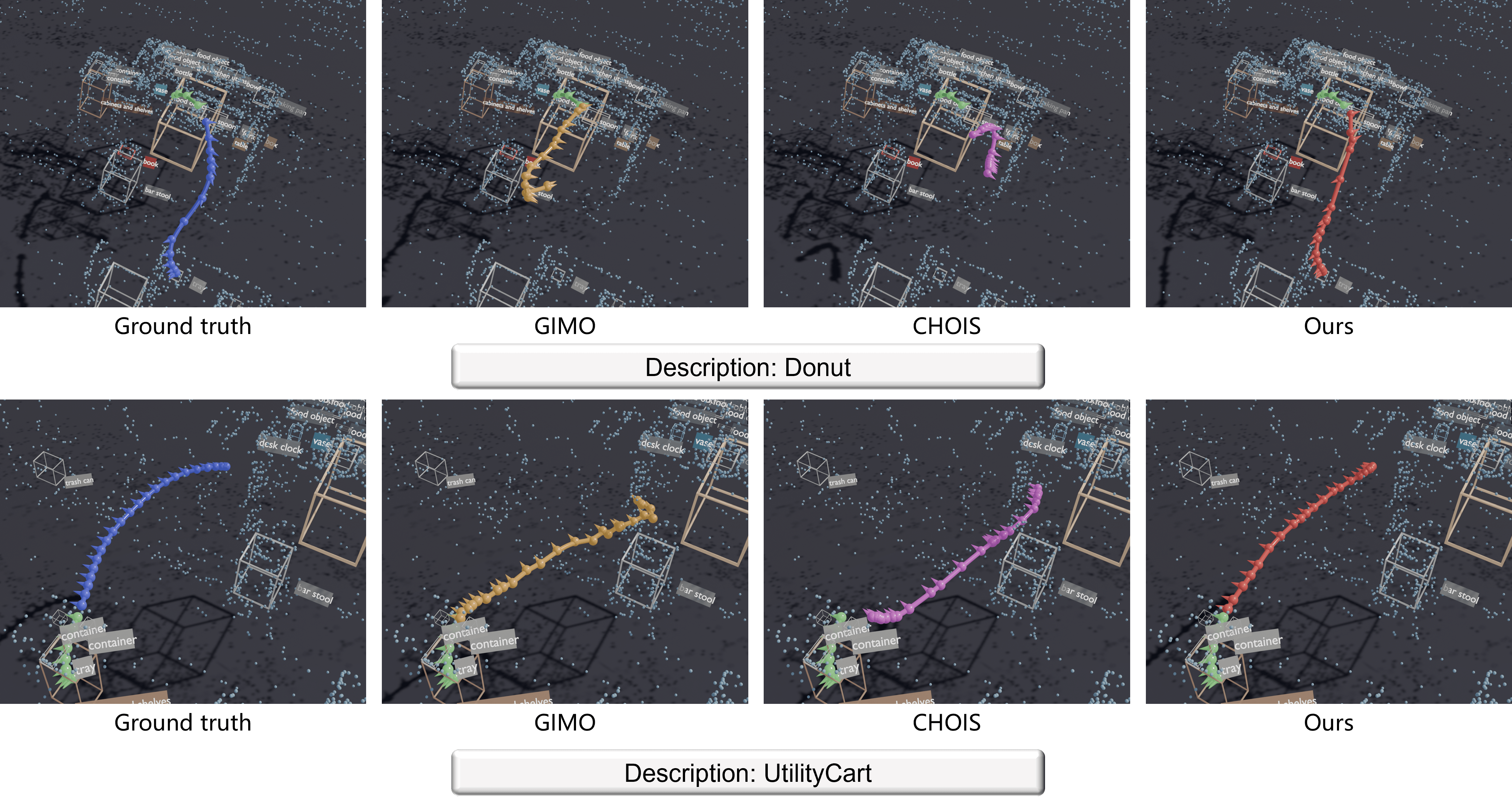}
    \caption{
    \textbf{Qualitative results in the ADT dataset.} 
    }
    \label{fig:supp_adt_ok}
\end{figure*}

\begin{figure*}[t]
    \centering
    \includegraphics[width=1\textwidth]{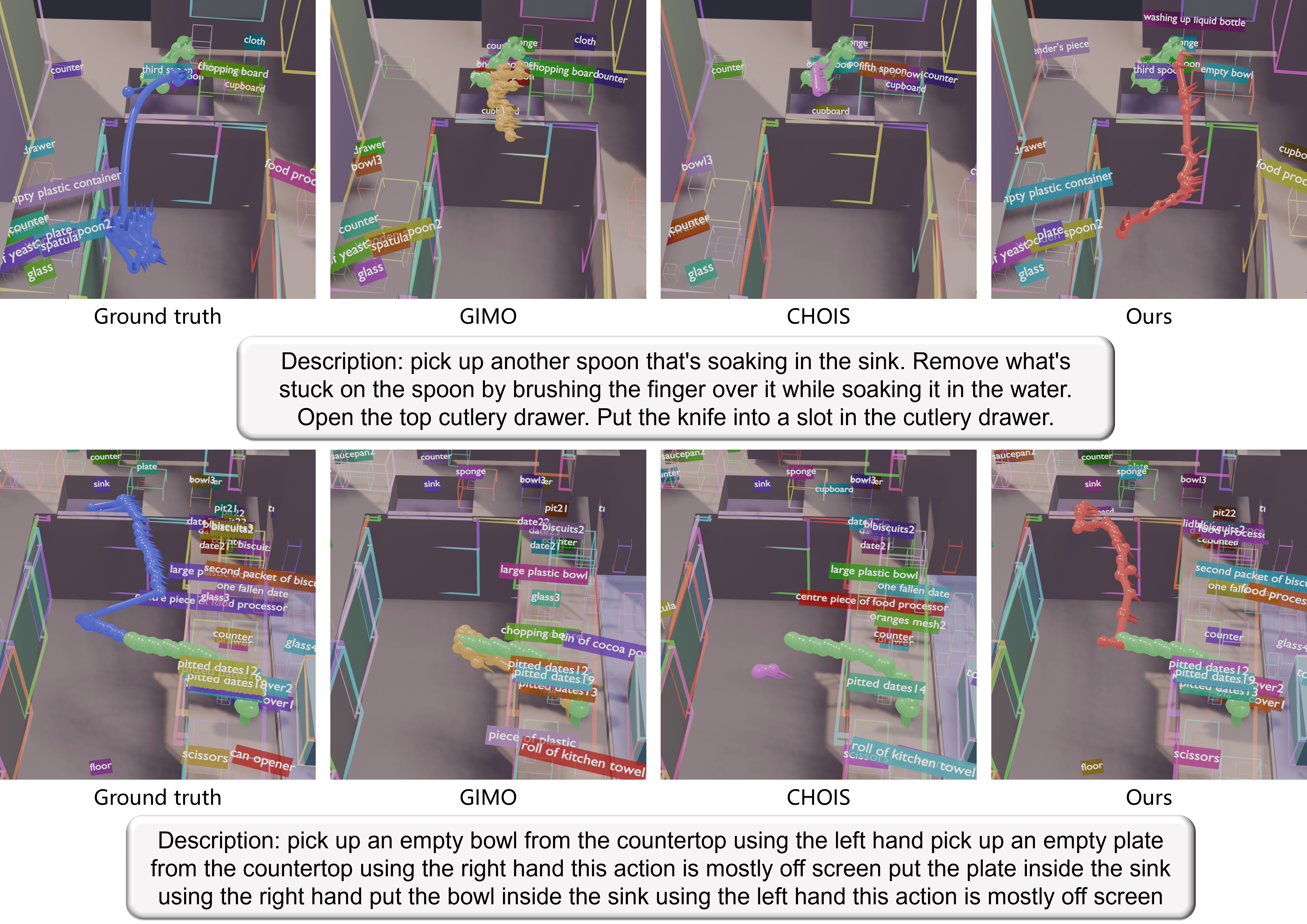}
    \caption{
    \textbf{Qualitative results in the HD-EPIC dataset.} 
    }
    \label{fig:supp_hdepic_ok}
\end{figure*}

\begin{figure*}[t]
    \centering
    \includegraphics[width=1\textwidth]{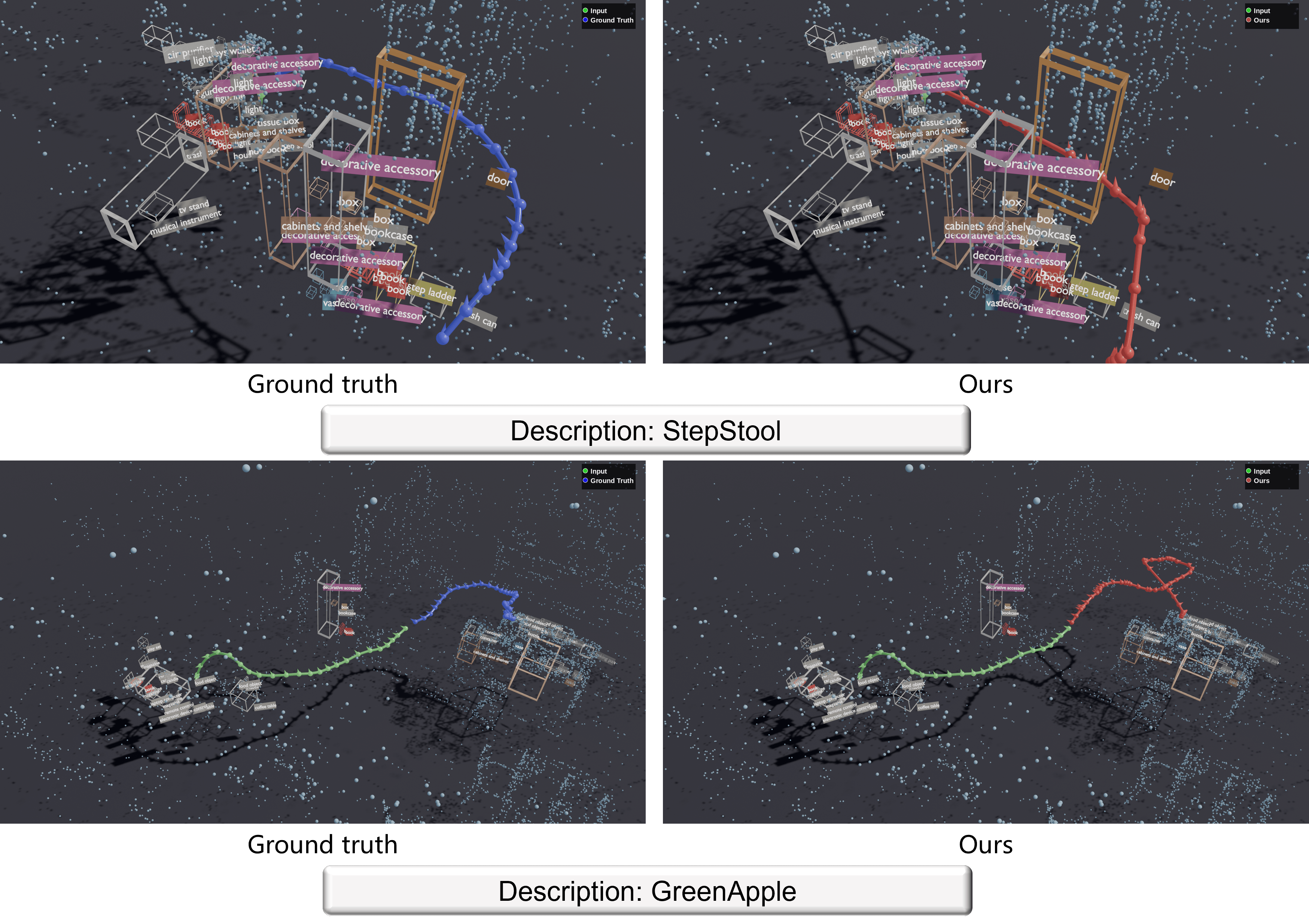}
    \caption{
    \textbf{Failure Cases in the ADT dataset.} 
    We can observe that sometimes, despite being goal-conditioned, the generated trajectory may be longer than the ground truth trajectory and may overshoot the destination. 
    }
    \label{fig:adt_failure}
\end{figure*}

\begin{figure*}[t]
    \centering
    \includegraphics[width=1\textwidth]{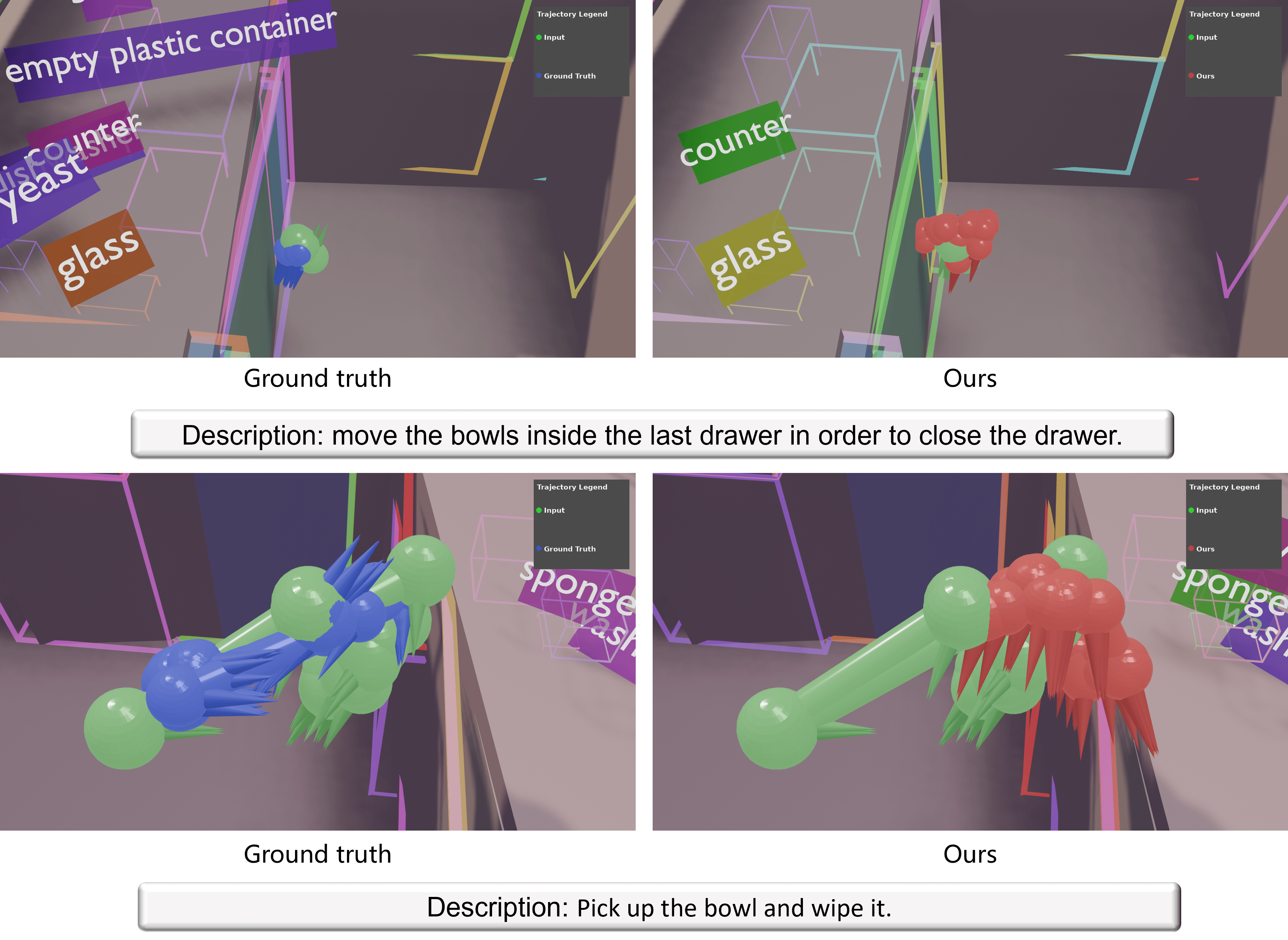}
    \caption{
    \textbf{Failure cases in the HD-EPIC dataset.} 
    Our method suffers from adding redundant motion for inputs don't have significant change in their positions. Such motion os observed for small object trajectories that are often interacted with for a very small duration.
    }
    \label{fig:hdepic_failure}
\end{figure*}

To complement the results presented in the main paper, we provide additional qualitative examples for both the ADT and HD-EPIC datasets. Figures~\ref{fig:adt_failure} and \ref{fig:hdepic_failure} illustrate representative failure cases and their likely causes, while Figs.~\ref{fig:supp_adt_ok} and \ref{fig:supp_hdepic_ok} present additional successful predictions.

